\documentclass[twoside,leqno,twocolumn]{article}

\usepackage[letterpaper]{geometry}
\usepackage{cuted}
\usepackage{ltexpprt}
\usepackage{hyperref}
\usepackage{adjustbox}
\usepackage{hyperref}
\usepackage{cite}
\usepackage{caption}
\usepackage{amsmath,amssymb,amsfonts}
\usepackage{algorithmic}
\usepackage{graphicx}
\usepackage{textcomp}
\usepackage{xcolor}
\newtheorem{definition}{Definition}
\def\BibTeX{{\rm B\kern-.05em{\sc i\kern-.025em b}\kern-.08em
    T\kern-.1667em\lower.7ex\hbox{E}\kern-.125emX}}
\begin{document}

\title{Non-Euclidean Spatial Graph Neural Network}

\author{Zheng Zhang\thanks{Emory University, Atlanta, Georgia, United States, \{zheng.zhang, sirui.li, allen.zhang, liang.zhao\}@emory.edu, junxiang.wang@alumni.emory.edu}
\and Sirui Li\footnotemark[1]
\and Jingcheng Zhou\thanks{The University of Manchester, Manchester, United Kingdom, j.zhou-74@sms.ed.ac.uk}
\and Junxiang Wang\footnotemark[1]
\and Abhinav Angirekula\thanks{Thomas Jefferson High School, Virginia, United States, abhinava2560@gmail.com}
\and Allen Zhang\footnotemark[1]
\and Liang Zhao\footnotemark[1]}
\date{}
\maketitle

%%
%% By default, the full list of authors will be used in the page
%% headers. Often, this list is too long, and will overlap
%% other information printed in the page headers. This command allows
%% the author to define a more concise list
%% of authors' names for this purpose.

%%
%% The abstract is a short summary of the work to be presented in the
%% article.
\begin{abstract}
    Spatial networks are networks whose graph topology is constrained by their embedded spatial space. Understanding the coupled spatial-graph properties is crucial for extracting powerful representations from spatial networks. Therefore, merely combining individual spatial and network representations cannot reveal the underlying interaction mechanism of spatial networks. Besides, existing spatial network representation learning methods can only consider networks embedded in Euclidean space, and can not well exploit the rich geometric information carried by irregular and non-uniform non-Euclidean space. In order to address this issue, in this paper we propose a novel generic framework to learn the representation of spatial networks that are embedded in non-Euclidean manifold space. Specifically, a novel message-passing-based neural network is proposed to combine graph topology and spatial geometry, where spatial geometry is extracted as messages on the edges. We theoretically guarantee that the learned representations are provably invariant to important symmetries such as rotation or translation, and simultaneously maintain sufficient ability in distinguishing different geometric structures. The strength of our proposed method is demonstrated through extensive experiments on both synthetic and real-world datasets.
\end{abstract}

%%
%% The code below is generated by the tool at http://dl.acm.org/ccs.cfm.
%% Please copy and paste the code instead of the example below.
%%

%\ccsdesc[500]{Computer systems organization~Embedded systems}
%\ccsdesc[300]{Computer systems organization~Redundancy}
%\ccsdesc{Computer systems organization~Robotics}
%\ccsdesc[100]{Networks~Network reliability}

%%
%% Keywords. The author(s) should pick words that accurately describe
%% the work being presented. Separate the keywords with commas.

%% A "teaser" image appears between the author and affiliation
%% information and the body of the document, and typically spans the
%% page.

%%
%% This command processes the author and affiliation and title
%% information and builds the first part of the formatted document.
\maketitle
\section{Introduction}
%Network data is a popular type of data that can describe the dependency information among data samples. In recent years, remarkable progress has been made towards representation learning on network data~\cite{hamilton2017representation,kipf2016semi,hamilton2017inductive}, and downstream applications such as recommender systems~\cite{ying2018graph}, drug discovery~\cite{gomez2018automatic,dai2018syntax,guo2020property, guo2021deep}, and natural language processing~\cite{marcheggiani2017encoding,bastings2017graph}. Specifically, the network data is usually described as a graph while the data samples in the network are denoted as nodes, and the edges are used to denote their dependencies. However, in many real-world networks, data sample connections are not only determined by their semantic dependencies but also constrained by the real-world spatial space in which they are embedded. Instead of simply viewing the network as a general graph by viewing coupling interactions as abstract edges between nodes, the embedded spatial space will significantly affect the interaction connections between data samples. Many previous studies named these networks spatial networks. 
\begin{figure}[t]
\centering
        \includegraphics[width=0.35\textwidth]{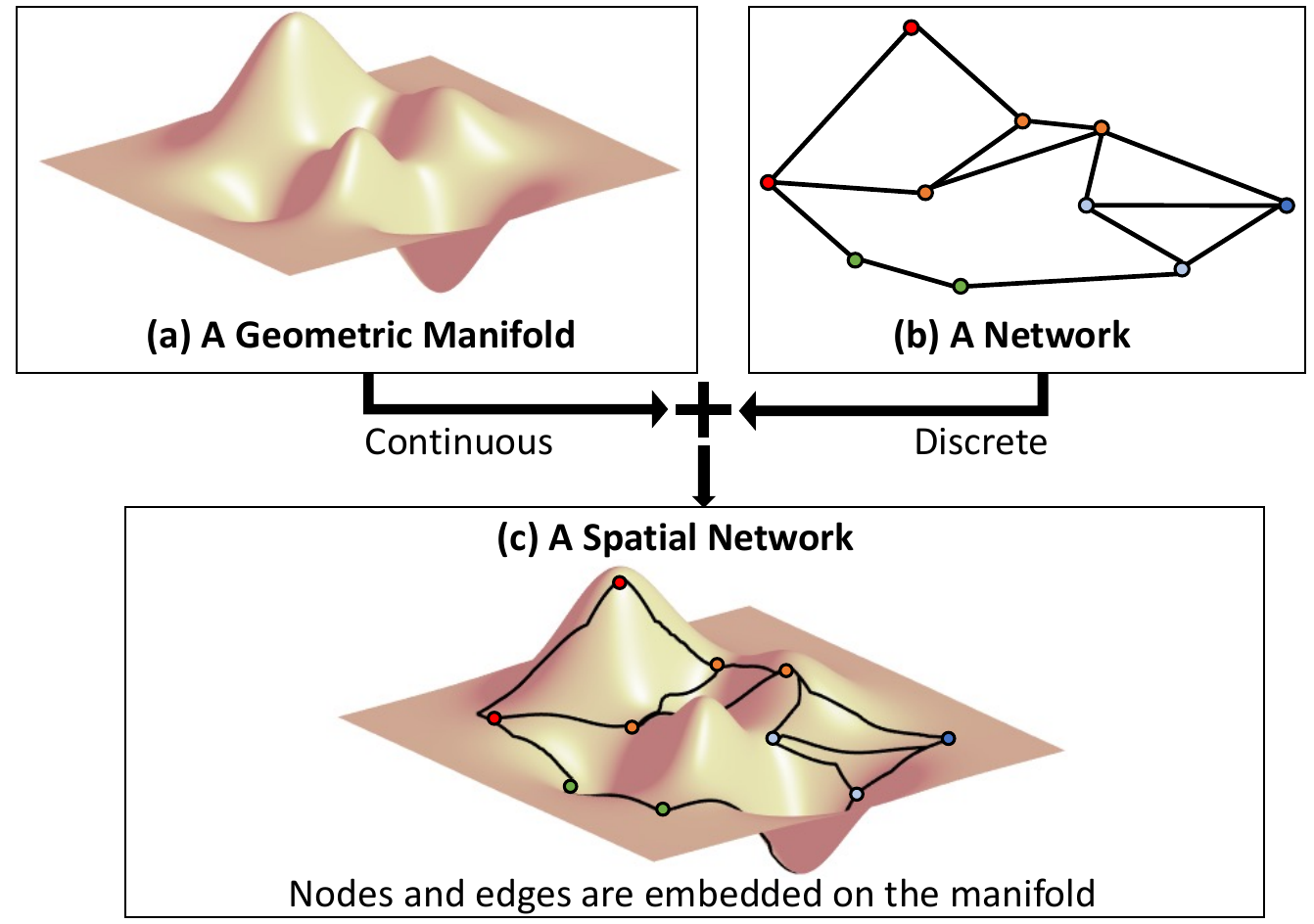}
        \vspace{-3mm}
        \captionsetup{width=1.0\linewidth}
        \caption{Spatial network contains not only network topology information but also their interaction with the embedded spatial surface.}
        \vspace{-7mm}
        \label{fig:intro} 
\end{figure}
Spatial networks are types of networks whose nodes and edges are embedded in a geometric spatial manifold. Spatial networks are ubiquitous in the real world, such as biological neural networks~\cite{eguiluz2005scale}, transportation networks~\cite{farahani2013review}, and mobility networks~\cite{chowell2003scaling}, where the spatial and network properties are deeply coupled together. For example, railroads are often built according to the terrain to achieve the lowest construction cost~\cite{salsbury2019emergence}. As shown in Figure~\ref{fig:intro}, understanding the mechanism of network interactions and their embedded spatial constraints, along with their coupled interaction, is crucial for learning effective representations of spatial networks. 

\begin{figure}[t]
\centering
        \includegraphics[width=0.38\textwidth]{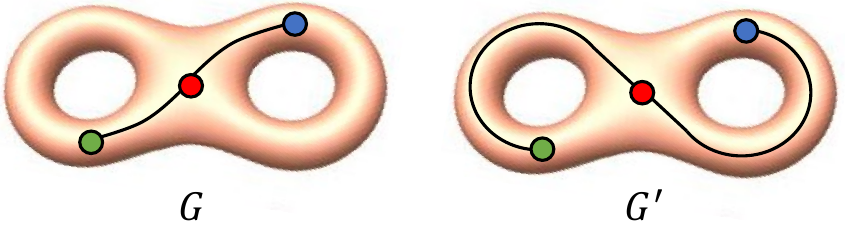}
        \vspace{-3mm}
        \captionsetup{width=1.0\linewidth}
        \caption{Two spatial networks with different connectivity mechanisms on a holomorphic manifold. The left figure reflects that nodes tend to be connected by the shortest distance (called the first law of geography~\cite{tobler1970computer}), while the right figure reflects the spatial pattern in which nodes tend to be connected by circuitous lines. Distinguishing these two spatial networks requires new approaches to jointly consider the spatial curves on the manifold and network topology.}
        \vspace{-7mm}
        \label{fig:example} 
\end{figure}

Although many efforts~\cite{erdHos1960evolution, watts1998collective, albert2002statistical, barabasi2003scale} have been put toward understanding the mechanism of spatial networks in some traditional research domains such as physics or mathematics, they usually require predefined human heuristics and prior knowledge of the analytical formulation of embedded spatial manifolds, which is usually unavailable in many real-world cases. In the era of deep learning, existing representation learning works on spatial networks~\cite{schutt2017schnet,klicpera2020directional,zhang2021representation} can only consider networks that are embedded in Euclidean space, where edge connections between nodes are described as straight lines. However, many real-world networks are embedded in non-Euclidean spaces, such as manifolds. The oversimplified approximations in flat Euclidean space will inevitably lose the rich geometric information carried by the irregular manifolds. Examples in Figure~\ref{fig:example} that share the same network topology and nodes' spatial coordinates, respectively, but with significantly different connecting curves between nodes, are non-distinguishable for existing representation learning methods on spatial networks. Therefore, jointly taking the irregularity of the embedded manifold with the network topology into account is crucial to extract powerful representations for spatial networks.

%it is usually more desirable to build a highway between two cities on a flag plain rather than a rugged mountain, since it is much more difficult and expensive to build a road on an irregular hill than on flat land.
%Learning effective representations of spatial networks is extremely challenging due to the coupled interactions between network and spatial topology information, and particular properties such as permutation invariant and rotation-translation invariant. 

%Existing works on spatial networks typically focus on learning the Euclidean space-approximated spatial networks. For example, by considering the spatial  However, in real-world datasets, the connections between nodes are typically embedded in a complicated surface with curved shapes and topology. Such over-simplified Euclidean space approximated connections can not consider the rich geometric information of the embedded surface, which results in the low expressive power of learned representations. 

%Besides considering the interactions of the network and its embedded spatial manifolds, .

%Simply summarizing the anomaly scores over nodes or edges can not reflect the anomalous degree of a subgraph in a collective way.
Despite the respective progress in representation learning on network data and spatial data respectively, the representation learning for spatial networks has been largely underexplored and has just started to attract fast-increasing attention, especially when considering the spatial space as an irregular non-Euclidean manifold. However, there is no trivial way to simply combine previous representation learning methods on network data and spatial data together to accomplish the task of representation learning on spatial networks due to several unique challenges: (1) \textbf{Difficulty in jointly considering discrete network and continuous spatial manifolds information, and their coupled interactions.} As shown in the example in Figure~\ref{fig:example}, some spatial networks may share the same spatial and network properties, respectively, but have significantly different interaction mechanisms. Simply combining spatial and graphical methods cannot distinguish these spatial networks.
(2) \textbf{Difficulty in extracting the geometric information of nodes and edges embedded in the irregular manifold.} In real-world situations, the manifolds that networks embed in are often irregular and inhomogeneous in space, where an explicit analytical form is usually infeasible. Thus, how to represent the geometric information of nodes and edges that are embedded in the manifold is challenging.
(3) \textbf{Difficulty for the learned representations to maintain rotation and translation invariance and all the geometric information.}
Rotation and translation invariance are natural and crucial for many applications of spatial networks~\cite{barthelemy2011spatial,hollard1982rotational}. How to learn rotation- and translation-invariant representations that still maintain sufficient ability in distinguishing different geometric patterns is challenging.

In order to design an effective method for learning powerful representations of spatial networks by addressing the above-mentioned challenges, we propose a novel method named \textbf{M}anifold \textbf{S}pace \textbf{G}raph \textbf{N}eural \textbf{N}etwork (\textbf{MSGNN}). To jointly learn network information and its embedded spatial manifold information, we propose a general learning message-passing framework. As shown in Figure~\ref{fig:framework}(a), in order to represent a continuous curve on a manifold, we first discretize the manifold into a mesh tessellation, and then learn the representation of curves through a sequential model of mesh units. Curve representations are then treated as messages on edges, and coupled spatial graph information is learned by passing and aggregating messages to nodes with graph convolutional layers.  As shown in Figure~\ref{fig:framework}(b), to deal with the irregularities of spatial curves and their embedded geometric manifolds, we propose to characterize several geometric features on each mesh unit of the curvilinear paths. Finally, we theoretically prove that the extracted spatial curve representations with a guarantee on the properties of \emph{rotation-invariant}, \emph{translation-invariant}, and \emph{geometric information-lossless}. To demonstrate the strength of our theoretical findings, extensive experiments are performed on both synthetic and real-world datasets.
\section{Related Work}
\noindent\textbf{Spatial Networks. }
There has been a long time of traditional research efforts on the subjects of spatial networks~\cite{barthelemy2011spatial}. Back to the field of quantitative geography forty years ago, \cite{haggett1969network} discovered the relevance of spatial constraints in the formation and evolution of networks, and developed models to characterize spatial networks. Advances in complex networks led to new insights regarding modern quantitative solutions~\cite{erdHos1960evolution}, appearing in more practical fields such as transportation networks~\cite{banavar1999size}, mobility networks~\cite{chowell2003scaling}, biological networks~\cite{eguiluz2005scale}, spatial social networks~\cite{johnson2003spatial,wang2022invertible}, and computational chemistry~\cite{gilmer2017neural}. However, existing works mostly focused on studying the Euclidean space networks and few attempts on non-Euclidean space require human domain knowledge about the analytical formulation of space, which is unrealistic for real-world cases.

\noindent\textbf{Geometric deep learning.}
In the deep learning era, many research works tried to extend standard deep learning methods~\cite{hamilton2017representation,kipf2016semi,hamilton2017inductive,zhang2021representation,du2021graphgt, ling2023deep,wang2022deep,zhang2023curriculum} to geometric data such as manifolds and graphs~\cite{cao2020comprehensive}. 
Existing deep learning methods on manifolds can be categorized according to the way they describe 3D data formats~\cite{wu2014shapenets,hang2015multi}. 3D ShapeNets\cite{wu2014shapenets} represents a geometric 3D shape as a probability distribution on a 3D voxel grid. Su et al.~\cite{hang2015multi} construct multi-view representations of 3D shapes by using multiple images from different viewpoints. The main disadvantage of these methods is the efficiency of representing 3D data. Models are difficult to scale because they require much higher complexity.
Alternatively, there exists another stream of works that represents manifold data by describing the 3D shape surface~\cite{litany2017deep}. Pioneering works~\cite{masci2015geodesic} generalize convolutional networks to non-Euclidean manifolds with a local geodesic system. Following works continue exploring the architecture of convolutions such as using localized spectral convolutional networks~\cite{li2016sync}. On the other hand, now there exist two significant streams of graph convolution. The first stream is spectral-based. Noticeable models such as ChebNet utilize spectral graph filters and their variants~\cite{tang2019cheb}. A main disadvantage of spectral approaches is that the spectral definition of convolution relies on the domain-dependent Fourier basis~\cite{cao2020comprehensive}. It is difficult to transfer a model learned on one graph to another graph with a different Fourier basis. The other stream of work is spatial-based, which directly performs convolution operations based in a message-passing manner. Methods such as GCN and GIN extract and aggregate neighborhood information for convolution~\cite{ niepert2016learning, xu2018powerful}.

\section{Problem Formulation}
%In this section, we first formalize spatial networks, and the problem of representation learning on spatial networks, then we introduce the challenges in order to solve this problem.

Spatial graphs (also known as spatial networks~\cite{barthelemy2011spatial}) are networks for which the nodes and edges are embedded in a geometric manifold surface. %{\ZZ {I first introduce the general manifold, then say the 3D situation. Is that too wordy? Shall I directly introduce the 3D situations?}} A $n$-\textit{manifold} is a Hausdorff topological space with a countable basis of open sets, such that each point of the manifold lies in an open set homeomorphic to $\mathbb{R}^n$. 
In this paper, we consider the connected smooth compact two-dimensional surface $M$, which is most commonly observed in our real-world 3D space. Locally around each point $x$ the manifold is homeomorphic to a two-dimensional Euclidean space referred to as the tangent plane and denoted by $T_xM$. Given the manifold $M$, a spatial network is typically defined as $G_M=(V,E,M_V,M_E)$ such that $V$ is the set of nodes and $E \subseteq V\times V$ is the set of edges. $e_{ij}\in E$ is an edge connecting nodes $v_i$ and $v_j\in V$. 
$M_V$ and $M_E$ denote the subset of the manifold $M$ that nodes and edges embed in, which is defined as $M_V\subseteq M, M_E \subseteq M$. Particularly, $M_V$ can be described as a set of 3D Cartesian coordinate points where we have $p_i \in M$ for each point $p_i$ representing the coordinates of node $v_i$. $M_E$ can be described as a set of curved lines that connect the nodes on the manifold, where for each $e_{ij}\in E$ we have its corresponding curved line as $l_{ij}\in M_E$. Specifically, a smooth curved line can be defined as a mapping function $l:[0,T]\rightarrow M_E$.

%To represent the continuous surface as discrete data format but still maintain the surface information, the manifold can be approximated by the format of mesh, which is denoted by the pair $M=(P,F)$, where $P=\{\mathbf{p}_1, \mathbf{p}_2,\dots\}$ is a set of 3D points that $\mathbf{p}_i=(x_i,y_i,z_i)|x_i,y_i,z_i \in \mathbb{R}$ in the Cartesian coordinate system, and $F$ is the set of faces that defines the connectivity (triplets of points for triangular meshes). Therefore, for each edge $e_{ij}\in E$ in the given network, it corresponds to a curved spatial line $l_{ij}$ that is embedded on manifold $M$, which corresponds to a set of faces $F^{(ij)} = \{f^{(ij)}_1, f^{(ij)}_2, \dots  f^{(ij)}_k| f^{(ij)}_t \in F\}$ that have overlapping with the line. In more detail, the path $l_{ij}$ is defined as a set of line segments $l^{(ij)} = \{l^{(ij)}_1, l^{(ij)}_2, \dots\, l^{(ij)}_k \}$, where each line segment $l^{(ij)}_k$ is embedded in its corresponding face $f^{(ij)}_k$.

The main goal of this paper is to learn the representation mapping function $f: G_M \rightarrow \mathbb{R}^{\mathcal{D}}$ to map an input spatial network to a high-dimensional vector, with the simultaneous satisfaction of strong discriminative power and significant symmetry properties, which is an extremely hard problem due to several unique challenges:
(1) {Difficulty in jointly considering discrete network and continuous spatial manifolds information, and their coupled interactions.} %As shown in the example in Figure xx, some spatial networks may share the same spatial and network properties, respectively, but have significantly different interaction mechanisms. Simply combining spatial and graphical methods cannot distinguish these spatial networks.
(2) {Difficulty in extracting the geometric information of nodes and edges embedded in the irregular manifold.} %In real-world situations, the manifolds that networks embed in are often irregular and inhomogeneous in space, where an explicit analytical form is usually infeasible. Thus, how to represent the geometric information of nodes and edges that are embedded in the manifold is challenging.
(3) {Difficulty for the learned representations to maintain rotation and translation invariance and all the geometric information.} The set of \textit{rotation and translation invariant functions} on the spatial network is defined as $f(G_M)=f(\mathcal{T}(G_{M}))=f((V,E,\mathcal{T}(M_V),\mathcal{T}(M_E)))$, for all $\mathcal{T} \in \mathrm{SE(3)}$, where $\mathrm{SE(3)}$ is the continuous Lie group of rotation and translation transformations in $\mathbb{R}^3$.
%Rotation and translation invariance are natural and crucial for many applications of spatial networks~\cite{barthelemy2011spatial,hollard1982rotational}.  How to learn rotation- and translation-invariant representations that still maintain a powerful ability in distinguishing different geometric patterns is challenging.
\vspace{-2mm}
\section{Method}%{\ZZ {Could you help me check this paragraph that summarizes our contributions}}

%In the rest of this section, we first introduce the details of the sequence representation of manifold-constrained spatial path in subsection~\ref{subsection:path-representation}. Then in subsection~\ref{subsection:Framework}, we present the design details of our whole framework to jointly combine network information and spatial path information. Finally, in subsection~\ref{subsection:theory}, we give the theoretical proof of the expressive power on learned representations and associated approximation bound.
\begin{figure*}[ht]
\centering
        \includegraphics[width=0.82\textwidth]{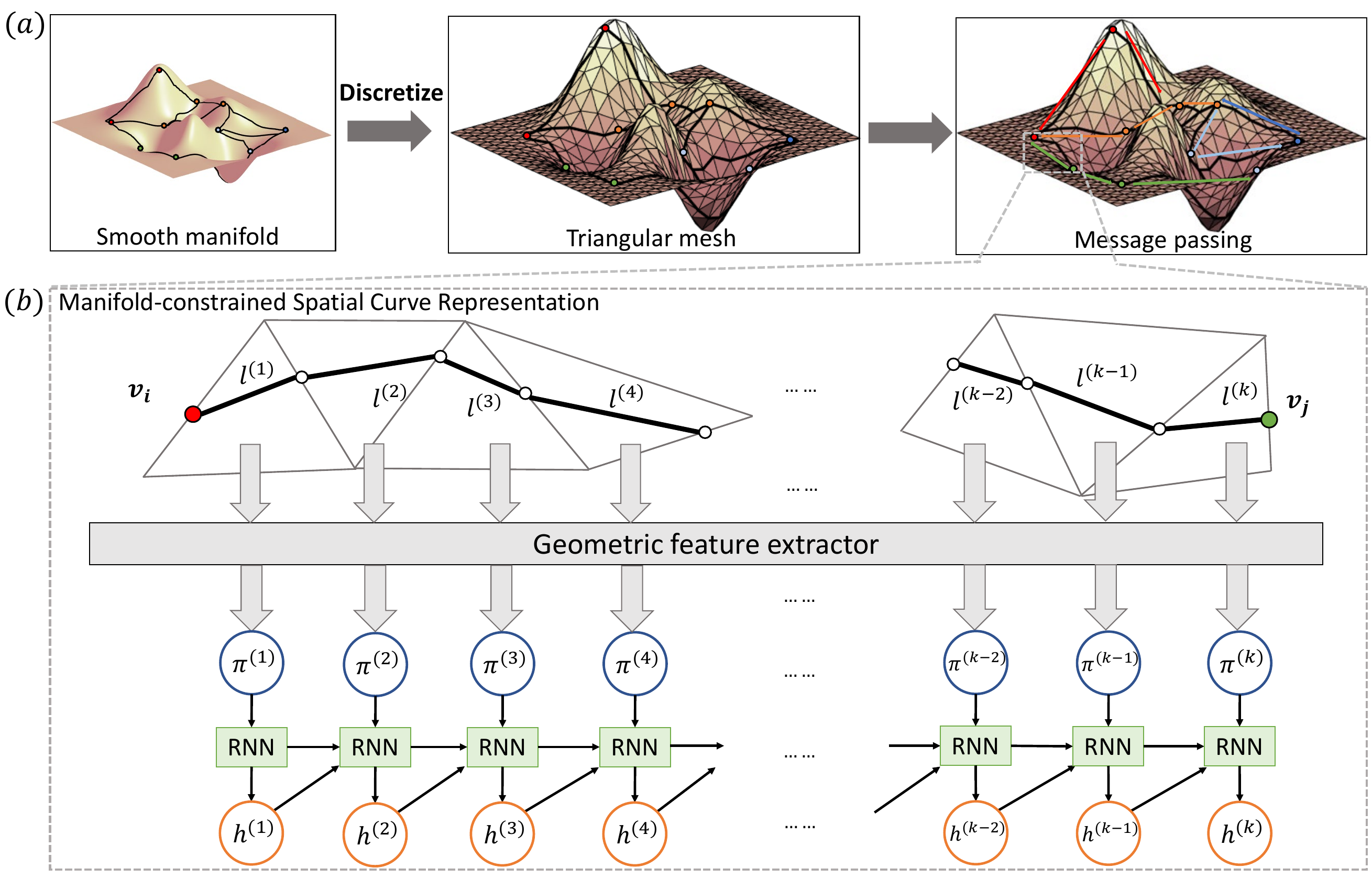}
        \vspace{-3mm}
        \captionsetup{width=1.0\linewidth}
        \caption{Illustration of the overall proposed framework. (a) The discretization process of the continuous manifold and convolutional neural networks for passing and aggregating the geometric information on spatial curves. (b) The RNN module extracts the geometric information along the irregular spatial curves between nodes. }
        \vspace{-5mm}
        \label{fig:framework} 
\end{figure*}

\subsection{Framework.} \label{subsection:Framework}
In this section, we first propose the overall framework that can jointly consider the network topology and its embedded spatial space information. Although there is a large amount of existing work on learning representations on network and spatial data separately, simply combining the network and spatial representations can not handle the coupled spatial graph information. Few existing works~\cite{schutt2017schnet,klicpera2020directional,zhang2021representation} on learning spatial networks representations only consider the spatial structure of nodes by considering their Euclidean coordinates. While for the edge connection between nodes, they simply treat it as a straight line connecting nodes in Euclidean space. Although these methods could achieve relatively good performance when the embedded space can be well approximated by Euclidean space, they can not well handle irregular and non-uniform embedding space, where the Euclidean space approximation will deviate largely from the real cases. Moreover, simply using Euclidean space approximation can not reflect the true geometry of the connection paths between nodes on the manifold surface, which will inevitably hurt the expressiveness of the learned representation. Thus, to fill this gap, in this paper we propose the first general framework to deal with spatial networks embedded in non-Euclidean manifold space. 

Due to the incompatibility between discrete network data and continuous spatial data, the first question is how to combine these two data in one end-to-end framework. Besides, the explicit analytical format function of the spatial manifold is usually extremely difficult to obtain because of the irregularity and non-uniformity of the real-world manifold surfaces. To address this problem, we first propose to discretize the continuous manifold space into discrete mesh data. Triangular mesh, which preserves shape surfaces and topology, is a popular format for efficiently approximating manifold shapes. Specifically, a triangular mesh can be defined as a collection of $C$ triangle faces $\mathcal{F}=\{f^{(1)},f^{(2)},\dots,f^{(C)}\}$, where the vertices of each $f^{(c)} \in \mathcal{F}$ are located on the surface of manifold $M$.

Given the discretized mesh to describe the embedded spatial surface, we can describe the spatial curved lines between nodes as a sequence of discrete units. As shown in Figure~\ref{fig:framework}(b), the sequence consists of triangular faces that the curve line passes. Therefore, the spatial information of each edge can be represented as a sequence of mesh units and line segments embedded on them. More concretely, for each edge $e_{ij}\in E$ exists in the given spatial graph, it corresponds to a curved spatial line $l_{ij}$ embedded on manifold $M$. Given the discretized triangular mesh, the embedded line $l_{ij}$ can be represented as a set of $K$ line segments $l_{ij} = \{l^{(1)}_{ij}, l^{(2)}_{ij}, \dots, l^{(K)}_{ij}\}$ with their corresponding embedded triangle faces $F_{ij} = \{f^{(1)}_{ij}, f^{(2)}_{ij}, \dots  f^{(K)}_{ij}\}$, where $K$ is the number of faces on this spatial path and each line segment $l^{(k)}_{ij}$ is embedded in its corresponding face $f^{(k)}_{ij}$. In summary, we can represent the spatial information on edge $e_{ij}$ as a sequence of pairs $((l^{(1)}_{ij},f^{(1)}_{ij}),(l^{(2)}_{ij},f^{(2)}_{ij}),\dots,(l^{(K)}_{ij},f^{(K)}_{ij}))$.

Given the sequences of pairs of faces and their embedded line segments, the next question is to incorporate them with the graph topology in a general model to learn the coupled spatial-graph representations. Since the length of the sequence may vary on different edges, some common methods such as MLP or CNN cannot be easily generalized to extract representations here. To address this issue, we propose a novel approach that mimics natural language processing approaches by analogizing each pair of spatial units as a token in a sentence, which is shown in Figure~\ref{fig:framework}(b). Therefore, a recurrent graph neural network (RNN) model such as GRU or LSTM is a natural choice to extract latent embeddings from these sequences. Formally, the extracted latent embedding $h(e_{ij})$ on edge $e_{ij}$ can be denoted as $\mathbf{RNN}(\pi(l^{(1)}_{ij},f^{(1)}_{ij}),\pi(l^{(2)}_{ij},f^{(2)}_{ij}),\dots,\pi(l^{(K)}_{ij},f^{(K)}_{ij}))$, where $\pi(\cdot)$ denotes the geometric information of the unit and will be introduced in Section~\ref{subsection:path-representation}. The extracted spatial information can then be treated as the message on graph edges. A graph message passing neural network model is then performed to jointly combine node information and all incoming messages on edges into updated node embeddings, where the update function is as follows:
\begin{align*}
\begin{split}
    \hat{h}(v_i) & = \mathbf{AGGREGATE}\{\mathbf{\xi}(h(v_j), h(e_{ij})) |j\in  \mathcal{N}(i)\},\\
    h(e_{ij}) & = \mathbf{RNN}(\pi(l^{(1)}_{ij},f^{(1)}_{ij}),\dots,\pi(l^{(K)}_{ij},f^{(K)}_{ij})),
\end{split}
\end{align*}
where $h$ represents latent embeddings and $\mathbf{\xi}$ denotes a nonlinear function such as multiple layer perceptron (MLP). $\mathbf{AGGREGATE}$ denotes any feasible set aggregate function.

%In order to jointly consider the discrete network and continuous spatial information, and their coupled interactions, here we propose a novel framework named XXX. 

\subsection{Manifold-constrained spatial curve representation.} \label{subsection:path-representation}
Given the above framework, a key question is how to define the spatial information extractor $\pi(\cdot)$ on each unit of a line segment and its embedded triangular mesh. As mentioned previously, we need a novel way to represent the spatial path information of the network edge that can preserve all the geometric shape information, and simultaneously maintain rotation- and translation-invariance. Obviously, simply feeding the Cartesian coordinates of each line segment in units can not guarantee invariance to the important symmetries such as rotation-and translation-transformations. Although there exist a few spatial information representation methods in the domain of spatial deep learning that can guarantee rotation and translation invariant features, we can not directly use them because they either can not capture the coupled spatial-graph properties because they are purely spatial-based methods, or can not guarantee all the geometric structure information is preserved because the information they extracted is not lossless. Here, the term lossless information means given the extracted geometric features, the information is sufficient to recover the input geometric structure. To handle this issue, for each sequence of spatial path line segments and their embedded meshes, we propose to extract a combination of geometric features on each mesh unit and the relative spatial relationship with their neighboring units. We theoretically guarantee the extracted information is sufficient to recover the full original geometries and also stay invariant to rotation and translation transformations. 

 %To guarantee invariance to rotation and translation transformations, we only compute geometry features with relative coordinates such as \textit{distance}, \textit{angle}, and \textit{torsion}, which will be shown as detailed in the following equations. 
Without loss of generality, we consider the spatial path between node $v_i$ and node $v_j$ in the given node set $V$ of graph that there exists an edge $e_{ij}$ between them. Formally, for each unit $(l^{(k)}_{ij},f^{(k)}_{ij})$ belongs to the sequence of edge $e_{ij}$, where $k\in[1,K]$, we use ${\pi}(l^{(k)}_{ij}, f^{(k)}_{ij})$ to represent the extracted geometric features on this unit. The sequence of spatial information on edge $e_{ij}$ is denoted as $((l^{(1)}_{ij},f^{(1)}_{ij}),(l^{(2)}_{ij},f^{(2)}_{ij}),\dots,$ $(l^{(K)}_{ij},f^{(K)}_{ij}))$. For the purpose of simplicity, we omit the subscript symbol $_{ij}$ in the rest of this section. 

To extract the necessary spatial information, we consider both the spatial information along the line segment and its embedded triangle mesh. The spatial information on the line segment, denoted as $\mathcal{\pi}(l^{(k)})$, and the embedded mesh face, denoted as $\mathcal{\pi}(f^{(k)})$, capture important geometric structure and relative orientation directions. For the line segment $l^{(k)}$, we extract its length $d^{(k)}$ and angle $\theta^{(k)}$ with respect to the connecting line $L$ between nodes $v_i$ and $v_j$. These features confine the relative position to a sphere in 3D space, allowing rotation around $L$. To fix the relative position, we further extract the torsion angles $\phi^{(k,k-1)}$ and $\phi^{(k,k+1)}$ between the current line segment $l^{(k)}$ and its neighboring segments $l^{(k-1)}$ and $l^{(k+1)}$.

For the embedded mesh faces $\mathcal{\pi}(f^{(k)})$, we consider the curvature vector direction to understand the spatial information of the surface environment. However, simply calculating the orientation $\mathbf{n}^{(k)}$ of the curvature vector does not guarantee rotation and translation invariance. To address this, we calculate the relative angles $\varphi^{(k,k-1)}$ and $\varphi^{(k,k+1)}$ between the curvature vectors of the given face and its neighboring faces. Additionally, we compute the angle $\varphi^{(k-1,k+1)}$ between the two neighboring curvature vectors. These angles form a triangle, ensuring fixed relative orientation.
Mathematically, the representation of spatial information for the $k$-th mesh unit on the spatial path sequence that forms the edge $e_{ij}$ between nodes $v_i$ and $v_j$ can be denoted as:
\begin{align}\label{eq:representation}
\begin{split}
   \pi(l^{(k)}, f^{(k)}) =& (d^{(k)}, \theta^{(k)}, \phi^{(k,k-1)}, \phi^{(k,k+1)},\\
   & \varphi^{(k,k-1)}, \varphi^{(k,k+1)}, \varphi^{(k-1,k+1)}),
\end{split}
\end{align}
where
\begin{align*}    \label{eq:representation_elaborate}
\begin{split}
      d^{(k)} & = ||\mathbf{l}^{(k)}||_2,       \theta^{(k)}  = \arccos{\langle \frac{\mathbf{l}^{(k)}}{d^{(k)}}, \frac{\mathbf{L}_{i,j}}{d_{i,j}} \rangle},  \\
       \quad \mathbf{L}_{i,j} & = \mathbf{p}_j - \mathbf{p}_i, d_{i,j}  = ||\mathbf{L}_{i,j}||_2 \\
      \phi^{(k,k-1)} & = \langle\frac{ \mathbf{c}^{(k-1)} \times \mathbf{c}^{(k)}}{||\mathbf{c}^{(k-1)} \times \mathbf{c}^{(k)}||_2},\frac{  \mathbf{L}_{ij}}{ ||\mathbf{L}_{ij}||_2}\rangle \cdot \bar{\phi}^{(k,k-1)}, \\
      \phi^{(k,k+1)} &=  \langle\frac{ \mathbf{c}^{(k)} \times \mathbf{c}^{(k+1)}}{||\mathbf{c}^{(k)} \times \mathbf{c}^{(k+1)}||_2},\frac{  \mathbf{L}_{ij}}{ ||\mathbf{L}_{ij}||_2}\rangle \cdot \bar{\phi}^{(k,k+1)}, \\
      \bar{\phi}^{(k,k-1)} & = \arccos\langle \mathbf{c}^{(k)} , \mathbf{c}^{(k-1)}\rangle,  \\
      \bar{\phi}^{(k,k+1)} & = \arccos\langle \mathbf{c}^{(k)} , \mathbf{c}^{(k+1)}\rangle ,\\
      \mathbf{c}^{(k)}   =&\frac{\mathbf{L}_{ij}\times \mathbf{l}^{(k)}}{||\mathbf{L}_{ij}\times \mathbf{l}^{(k)}||_2}, \varphi^{(k-1,k+1)}  = {\arccos{\langle \mathbf{n}^{(k-1)} ,\mathbf{n}^{(k+1)} \rangle}} , \\
      \mathbf{c}^{(k-1)}   =&\frac{\mathbf{L}_{ij}\times \mathbf{l}^{(k-1)}}{||\mathbf{L}_{ij}\times \mathbf{l}^{(k-1)}||_2}, \varphi^{(k-1,k)}  = {\arccos{\langle \mathbf{n}^{(k-1)} ,\mathbf{n}^{(k)} \rangle}}, \\
      \mathbf{c}^{(k+1)}   =&\frac{\mathbf{L}_{ij}\times \mathbf{l}^{(k+1)}}{||\mathbf{L}_{ij}\times \mathbf{l}^{(k+1)}||_2}, \varphi^{(k,k+1)}  = {\arccos{\langle \mathbf{n}^{(k)} ,\mathbf{n}^{(k+1)} \rangle}}     
\end{split} 
\end{align*}
\begin{theorem}   
Here the distances $d \in [0,\infty)$, angle $\theta \in [0, \pi)$, torsions $\phi\in [-\pi, \pi)$, and relative orientation angle $\varphi\in [0, \pi)$ are rigorously invariant under all rotation and translation transformations $\mathcal{T}\in \mathrm{SE(3)}$.
\label{theo:theorem1}
\end{theorem}

The proof is straightforward and can be found in Appendix~\ref{appendix:theorem1}. 
Intuitively, distance, angle, torsion, and orientation angle are invariant to translation and rotation transformations, since only relative coordinates are used in the formula. 
It is remarkable to mention that the proposed representation in Equation~\ref{eq:representation} not only satisfies the invariance under rotation and translation transformation but also retains the necessary information to recover the entire geometric structure of original spatial networks under weak conditions, as described in the following theorem.
\begin{theorem}
Given a spatial network $G_M=(V,E,M_V,M_E)$, if $G_M$ is a connected graph, then for any edge $e_{ij}$ with spatial curve sequence that has length of sequence $\tau\geq 3$, given Cartesian coordinates of two endpoints and one arbitrary point, the whole Cartesian coordinates of the given spatial networks can be determined by the spatial representation defined in Equation~\ref{eq:representation}.
\label{theo:theorem2}
\end{theorem}
The detailed proofs can be found in Appendix~\ref{appendix:theorem2}.

In addition, we further theoretically prove that the approximation error bound of discretized mesh representation in the Appendix \ref{subsection:theory}.

\subsection{Complexity analysis}
The time complexity of an $L$-layer GNN is $O(L\left|\mathcal{E}\right|b+L\left|\mathcal{V}\right|b^2)$, where $b$ is the number of latent dimensions. Second, the time complexity of extracting geometric features from edge trajectory by LSTM is $O(L\left|\mathcal{E}\right|(Kb^2+Kbd))$ where $K$ is the average length of spatial trajectories and $d$ is the number of computed geometric features. Therefore, the total time complexity of our algorithm is $O(L\left|\mathcal{E}\right|(Kb^2+Kbd+b)+L\left|\mathcal{V}\right|b^2)$.
%\begin{sproof}This can be achieved by Lemma 2 and Lemma 5 by \cite{khoury2019approximation} and the property that angles satisfy the triangle inequality.\end{sproof}
\indent The complete proof of Theorem \ref{theo: angle bound} is shown in the Appendix due to space limit. Theorem \ref{theo: angle bound} shows that the angle bound is defined by the plane angle and the circumradius.

\section{Experimental Results}
In this section, we first introduce the experimental settings, then the effectiveness of our proposed framework on both synthetic and real-world datasets is presented. The link to our code is at the GitHub repository \url{https://github.com/rollingstonezz/SDM24_Manifold_spatial_networks}.
\vspace{-3mm}
\subsection{Experimental Settings.} 
\subsubsection{Synthetic datasets.} In order to examine the effectiveness of our proposed MSGNN method in learning the coupled network and spatial manifold information, we follow previous works~\cite{barthelemy2011spatial} to generate a set of synthetic datasets. We generalize the preferential attachment model~\cite{albert2002statistical} to a spatial variant that all nodes and edges are embedded in a defined spatial manifold surface. Specifically, we first randomly generate a manifold surface in 3D space from a designed candidate pool of geometric shapes such as sphere or paraboloid. The process to generate such spatial networks starts from an initial connected network of $m_0$ nodes that are randomly sampled on the manifold surface. Then we introduce a new node $v_j$ to connect to the existing network at each iteration step. The new node is allowed to make $m\leq m_0$ connections towards existing nodes with a probability $\Pi_{j\rightarrow i}\sim k_i F[d_g(i,j)]$, where $k_i$ is the degree of node $v_i$ and $F$ is an exponential function $F(d_g)=e^{-d_g/r_c}$ of the geodesic distance $d_g(i,j)$ between the newly added node $v_j$ and the node $v_i$ on the manifold. Therefore, the formation mechanism of generated spatial networks is jointly determined by the spatial and network information. General characteristics of spatial networks~\cite{barthelemy2011spatial} such as spatial diameter $D$, and spatial radius $r$ are set as the prediction targets. Besides, we also add the interaction range $r_c$, which is a significant coupled spatial-graph label that affects the formation of the spatial networks, as another prediction target. We vary the type of embedded manifolds and other parameters of spatial networks to collect $5,000$ samples.% in our synthetic dataset.

\begin{table*}[t]
\begin{adjustbox}{width=.85\linewidth,center}
\begin{tabular}{c|ccccccccc}
\hline
Target & GCN  & GIN        & PointNet   & PPFNet     & MeshCNN    & CurvaNet   & SchNet     & SGMP       & MSGNN       \\ \hline
$r_c$ & 3.35$\pm$0.14 & 2.55$\pm$0.18 & 2.45$\pm$0.08 & 2.68$\pm$0.11 & 2.06$\pm$0.13 & 1.54$\pm$0.10 & \underline{0.97$\pm$0.06} & 1.08$\pm$0.05 & \textbf{0.83$\pm$0.04} \\
$D$   & 2.20$\pm$0.15 & 2.73$\pm$0.21 & 1.82$\pm$0.06 & 1.98$\pm$0.12 & 1.87$\pm$0.06 & 1.93$\pm$0.11 & 1.94$\pm$0.10 & \underline{1.86$\pm$0.05} & \textbf{1.67$\pm$0.05} \\
$r$   & 2.63$\pm$0.14 & 2.60$\pm$0.26 & 1.95$\pm$0.09 & 2.07$\pm$0.09 & 1.60$\pm$0.07 & 1.74$\pm$0.06 & 1.98$\pm$0.07 & \underline{1.88$\pm$0.05} & \textbf{1.47$\pm$0.11} \\ \hline
\end{tabular}
\end{adjustbox}
    \vspace{-3mm}
    \captionsetup{width=0.98\linewidth}
    \caption{The RMSE results of the synthetic dataset. The best performance for each predictive target is shown in bold, while we also underline the second-best performing models.}
    \label{table:synthetic_results} 
    \vspace{-3mm}
\end{table*}

\subsubsection{Real-world datasets} 
To further evaluate the performance of our proposed MSGNN and comparison methods in real-world scenarios, five public benchmark real-world spatial network datasets with different application domains are utilized as benchmark datasets in our experiments. Specifically, we include one brain network dataset and two 3D shapes datasets for graph classification task, and two airline transportation networks for link prediction task. We provide a brief description of these datasets as follow and detailed introduction can be found in Appendix~\ref{appendix:dataset}.

\noindent (1) \textbf{HCP brain networks.} 
Classify the activity states of subjects based on processed functional connectivity (FC) networks derived from the human brain manifold environment. Each data sample is associated with an activity state (e.g., rest, gamble) as the target for prediction. To construct brain networks, a threshold is applied to the FC values to filter out highly correlated edges. The spatial trajectories of interest are defined as geodesic paths between the centers of the ROIs.

\noindent (2) \textbf{Air transportation networks.} We adopt two publicly available flight networks \textit{Flight-NA} and \textit{Flight-GL}, where \textit{Flight-NA} contains 456 airports and 71,959 airlines in the North America and \textit{Flight-GL} contains 3,214 airports and 66,771 airlines spanning the globe. The earth surface is considered as the manifold to include the curved airline trajectory.

\noindent (3) \textbf{3D shapes classification.} We further conduct experiments on classifying 3D shapes in two datasets \textit{SHREC} and \textit{FAUST}. We follow previous studies~\cite{hanocka2019meshcnn} to sample a lower resolution ($\sim$500 faces) from a higher solution. The vertices of the lower-resolution triangle tessellation are then treated as the nodes and their geodesic trajectories are treated as the spatial curves.

\begin{table*}[t]
\begin{adjustbox}{width=1.0\linewidth,center}
\begin{tabular}{c|ccccccccc}\hline
          & GCN         & GIN         & PointNet    & PPFNet      & MeshCNN     & CurvaNet    & SchNet      & SGMP        & MSGNN        \\\hline
HCP       & 0.835$\pm$0.014 & 0.920$\pm$0.007 & 0.845$\pm$0.027 & 0.876$\pm$0.008 & 0.784$\pm$0.031 & 0.759$\pm$0.025 & 0.896$\pm$0.012 & \underline{0.927$\pm$0.004} & \textbf{0.951$\pm$0.005} \\
Flight-NA & 0.674$\pm$0.015 & 0.706$\pm$0.006 & 0.694$\pm$0.011 & 0.698$\pm$0.004 & 0.521$\pm$0.023 & 0.597$\pm$0.019 & 0.710$\pm$0.008 & \underline{0.719$\pm$0.004} & \textbf{0.730$\pm$0.005} \\
Flight-GL & 0.722$\pm$0.003 & 0.756$\pm$0.014 & 0.737$\pm$0.010 & 0.715$\pm$0.012 & 0.556$\pm$0.032 & 0.628$\pm$0.025 & 0.750$\pm$0.008 & \underline{0.761$\pm$0.009} & \textbf{0.785$\pm$0.005} \\
SHREC     & 0.525$\pm$0.042 & 0.533$\pm$0.034 & 0.567$\pm$0.007 & 0.887$\pm$0.010 & \underline{0.910$\pm$0.003} & 0.902$\pm$0.004 & 0.575$\pm$0.012 & 0.896$\pm$0.005 & \textbf{0.918$\pm$0.004} \\
FAUST     & 0.535$\pm$0.010 & 0.783$\pm$0.013 & 0.905$\pm$0.010 & 0.918$\pm$0.005 & 0.903$\pm$0.008 & \underline{0.923$\pm$0.004} & 0.865$\pm$0.023 & 0.840$\pm$0.035 & \textbf{0.925$\pm$0.005} \\\hline
\end{tabular}
\end{adjustbox}
\vspace{-3mm}
    \captionsetup{width=0.98\linewidth}
    \caption{The accuracy results of the real-world datasets. The best performance for each predictive target is shown in bold, while we also underline the second-best performing models.}
    \label{table:realworld_results} 
    \vspace{-5mm}
\end{table*}
\subsubsection{Comparison models}
To the best of our knowledge, there is little previous work that can handle general spatial networks.
We compare our proposed MSGNN against several categories of competitive methods, spanning two graph neural networks methods \textbf{GCN}~\cite{kipf2016semi} and \textbf{GIN}~\cite{xu2018powerful}, two spatial deep learning methods on point clouds \textbf{PointNet}~\cite{qi2017pointnet} and \textbf{PPFNet}~\cite{deng2018ppfnet}, two spatial deep learning methods on mesh \textbf{MeshCNN}~\cite{hanocka2019meshcnn} and \textbf{CurvaNet}~\cite{he2020curvanet}. We also include two state-of-the-art deep learning methods on Euclidean spatial networks \textbf{SchNet}~\cite{schutt2017schnet} and \textbf{SGMP}~\cite{zhang2021representation}. To ensure a fair comparison, we provided Cartesian coordinates as node attributes for GNN-based methods. For spatial deep learning methods on point clouds and mesh, we augmented the node attributes with graph connectivity information obtained from a trained Node2Vec model~\cite{grover2016node2vec}. Additionally, we established a consistent search range for model hyperparameters, such as the number of convolutional layers or the dimensionality of hidden embeddings, to maintain fairness across all models.

\begin{figure}[t]
    \centering
        \includegraphics[width=0.5\textwidth]{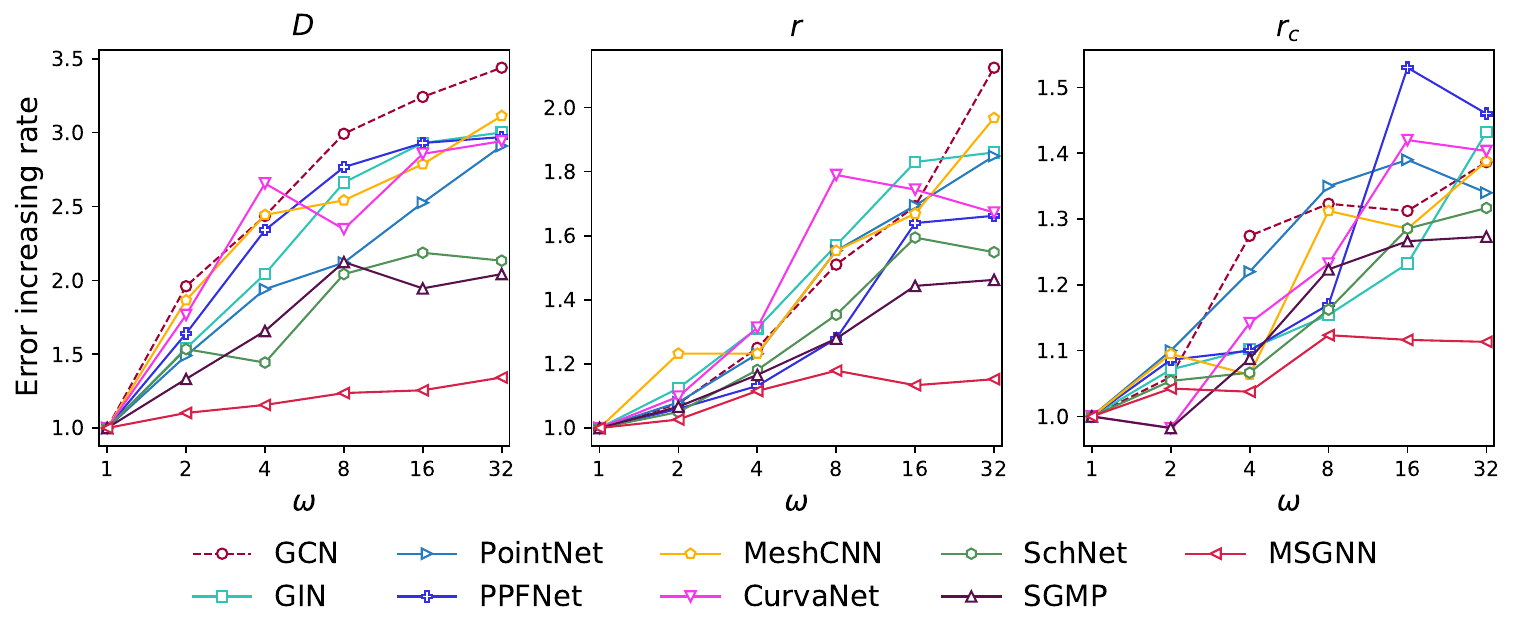}    
        %\captionsetup{font=small}
        \vspace{-8mm}
        \captionsetup{width=0.98\linewidth}
        \caption{Accuracy trend results for our proposed MSGNN model and all competing models against varying degree of manifold irregularity. The performance of models at the lowest degree of irregularity ($\omega=1$) is set as the base value.  }
        \label{fig:syn_error_increase} 
        \vspace{-8mm}
\end{figure}
\subsection{Effectiveness Results}
\vspace{-2mm}
\subsubsection{Synthetic datasets results.}
\vspace{-3mm}
Here we report the root mean squared error (RMSE) results of our proposed MSGNN with comparison methods on the synthetic dataset in Table~\ref{table:synthetic_results}. We summarize our observations on the model effectiveness below:

\noindent (1) The results demonstrate the strength of our proposed MSGNN by consistently achieving the best results in predicting all three coupled spatial-graph targets. Specifically, our model outperformed all the benchmark models by over $38.0\%$ on average, as well as outperformed the second-best model by $17.6\%$ on average. 

\noindent (2) Our proposed MSGNN method consistently achieves superior performance with respect to all predictive targets, which proves the robustness of MSGNN. In comparison, the spatial neural network methods on point clouds and mesh have significantly different performances on different tasks. For example, they shows competitive performance to spatial network methods on targets spatial diameter $D$, and spatial radius $r$. While their performance on predicting interaction range $r_c$ is significantly worse than spatial network methods by over $53.0\%$ on average, which may indicate that simply combining graph and spatial representation can not capture the interactions between these two data sources.

\noindent (3) It is also worth noting that the category of methods on spatial networks (SchNet, SGMP, and MSGNN) show a more competitive performance than methods in other categories, by over $34.2\%$ on average, which indicates that either graph neural network or spatial neural network methods have limited capability to effectively learn coupled spatial-graph properties. MSGNN shows a stronger performance compared to other methods on spatial networks by  $18.2\%$ on average, which demonstrates our method takes advantage of the irregular manifold information within the context of the spatial paths to acquire a more competitive performance.

\subsubsection{Real-world datasets results.}
Here we report the accuracy results of our proposed MSGNN with comparison methods on the real-world dataset in Table~\ref{table:realworld_results}. 

\noindent (1) Our proposed MSGNN method consistently achieved the best results among all methods in all five real-world datasets. Specifically, our results outperformed all the benchmark models by over $14.1\%$ on average and outperformed the second-best model by $4.2\%$ on average. The superior performance demonstrates the effectiveness of MSGNN for learning powerful representations in complex real-world scenarios.

\noindent (2) In two air transportation networks, our method achieves more considerable performance gains on the global network (Flight-GL) than on the North American network (Flight-NA). One possible reason is that North America is relatively small compared to the globe, and the curved effect on the surface is not significant. This may indicate that our method can exploit the curvature of the embedded surface to further improve the representation ability.

\noindent (3) Different classes of methods perform significantly differently on different datasets. For example, the class of spatial neural networks on mesh (MeshCNN and CurvaNet) have achieved competitive results in 3D shapes classification tasks (SHREC and FAUST) by outperforming other benchmark models by $13.1\%$ on average. However, they also performed poorly on air transportation and brain datasets by achieving the worst performance among all classes of methods. Such behavior indicates that these methods can not well handle generic spatial networks.

\noindent (4) It is also worth noting that on the SHREC dataset, only methods that consider orientation information achieved competitive results, which may arguably indicate that orientation information is important in the prediction task on this dataset.

Our method's invariance to rotation and translation is detailed in Appendix~\ref{appendix:invariant} due to space constraints.
\subsection{Effect of manifold irregularity analysis.}
Compared to existing representation learning methods on spatial networks, a contribution of our work is that our method can handle irregular geometric manifolds, rather than simply using Euclidean space approximations. To investigate the impact of manifold irregularity on model performance, we further introduce a series of experiments to vary the degree of irregularity of the manifold embedded by the network. Specifically, in our synthetic dataset setting, we choose a sinusoidal surface as the manifold for embedding the network, and we use a frequency parameter $\omega$ to control the irregularity of the generated manifold surface. The mathematical formulation of the sinusoidal surface can be written as $z=\sin(\omega  \sqrt{x^2+y^2})$. Larger values of $\omega$ here indicate that the resulting manifold surface will have a larger degree of irregularity. We vary the value of $\omega$ from $1$ to $32$ to generate a total of 6 datasets, and we compare the performance of predicting three targets on our method and competing methods. Particularly, we set the RMSE performance of all models at $\omega=1$ as a benchmark, and then calculate the increasing rate of RMSE when $\omega$ increases. The results are shown in Figure \ref{fig:syn_error_increase}.
According to the figure, compared to all baseline models, our model consistently shows a significantly slower increasing trend of the RMSE error as the degree of manifold irregularity increases. Such model behavior demonstrates that our model can effectively handle the irregularity in geometric manifolds. More interestingly, our model also shows a convergence trend in predicting the spatial radius $r$ and the interaction range $r_c$, which arguably further demonstrates the robustness of our model to extremely irregular manifold environments.

\begin{table}[t]
\begin{adjustbox}{width=1.0\linewidth,center}
    \begin{tabular}{l|cccccc}
    \hline
    Number of faces & 1,000                       & 2,000                      & 4,000                  &   8,000                     & 16,000             & 32,000             \\ \hline
    HCP       & 0.928 & 0.939 & 0.944 & 0.947 & 0.950 & 0.951 \\  
    Flight-GL       & 0.773 & 0.780 & 0.782 & 0.784 & 0.785 & 0.785 \\ \hline
    \end{tabular}
\end{adjustbox}
    \captionsetup{width=0.98\linewidth}
    \caption{Sensitivity analysis of model performance against the number of discretized mesh units.}
    \label{table:num_mesh} 
    \vspace{-5mm}
\end{table}

\subsection{Sensitivity analysis.}
We investigate the impact of the number of triangle mesh tessellations on our method to test the sensitivity of our model. We vary the number of mesh faces from high-resolution $32,000$ to low-resolution $1,000$ on two real-world datasets as shown in Table \ref{table:num_mesh}
\noindent (1) According to Table \ref{table:num_mesh}, with the increase of triangular mesh subdivision number, the accuracy scores on all datasets show a upward and converging trend. The convergent performance trends demonstrate the effectiveness of using mesh tessellations to approximate spatial manifold surfaces. Specifically, as the resolution of the mesh tessellations increases, the approximate discrete surface is approaching the underlying continuous manifold space. 
\noindent (2) It is also worth noting that as the number of mesh tessellations decreases, the performance of our proposed MSGNN on all methods gradually approaches and converges to the spatial network method on Euclidean space. The reason is that as the resolution of the mesh subdivision decreases, the approximate spatial path between nodes eventually converges to a Euclidean approximation. 

%In this subsection, we investigate the impact of parameters on the effectiveness of our proposed method by a series of experiments.

\vspace{-2mm}\section{Conclusion}
This paper focuses on the crucial problem of learning representations from spatial networks embedded in non-Euclidean manifolds, which is an underexplored area and can not be well handled by existing works. The proposed framework \textbf{M}anifold \textbf{S}pace \textbf{G}raph \textbf{N}eural \textbf{N}etwork (\textbf{MSGNN}) effectively addresses the unique challenges of representing irregular spatial networks by first converting the manifold space into a discrete mesh tessellation, and then converting the geometric information of the curves between nodes into messages on edges. Theoretical guarantees are given to prove that our learned representations are invariant to important symmetries such as rotation and translation, and simultaneously maintain strong distinguish power in geometric structures. Extensive experimental results on both synthetic and real-world datasets demonstrate the strength of our theoretical findings.

%%
%% The next two lines define the bibliography style to be used, and
%% the bibliography file.
\bibliographystyle{plain}
\bibliography{ref}

%%
%% If your work has an appendix, this is the place to put it.
\clearpage
\appendix

\section{Mathematical Proofs}

\subsection{Proof of Theorem \ref{theo:theorem1}}\label{appendix:theorem1}
\begin{proof}
Intuitively, distance, angle, torsion, and orientation angle are invariant to translation and rotation transformations, since only relative coordinates are used in the formula. 
Formally, for translation transformations $\mathcal{T}\in SE(3)$ and rotation transformations $\mathcal{R}\in SE(3)$, the following identity equations hold:
\begin{equation}
\begin{split}
\mathcal{T}(\mathbf{x}-\mathbf{y}) &= \mathbf{x}-\mathbf{y},  \\
\langle \mathcal{R}(\mathbf{x}), \mathcal{R}(\mathbf{y}) \rangle  &  = \langle \mathbf{x}, \mathbf{y} \rangle \\
 \mathcal{R}(\mathbf{x}) \times \mathcal{R}(\mathbf{y}) & = \mathcal{R}(\mathbf{x} \times\mathbf{y}) 
\end{split}
\label{eq:app_repre_1}
\end{equation}
Thus we have 
\begin{align*}    
\begin{split}
      d & = ||\mathbf{l}||_2 = ||\mathbf{p}-\mathbf{p}^{\prime}||_2 = ||\mathcal{T}(\mathbf{p})-\mathcal{T}(\mathbf{p}^{\prime})||_2,     \\  
      d & = ||\mathbf{l}||_2 = \langle \mathbf{l}, \mathbf{l} \rangle = \langle \mathcal{R}(\mathbf{l}), \mathcal{R}(\mathbf{l}) \rangle ,  \\
      \theta & = \arccos({\langle \frac{\mathbf{l}}{d}, \frac{\mathbf{L_{ij}}}{d} \rangle})   \\
      & = \arccos({\langle \frac{\mathcal{R}(\mathbf{l})}{d}, \frac{\mathcal{R}(\mathbf{L_{ij}})}{d} \rangle}),\\
      \phi^{(k,k+1)} &=  \langle\frac{ \mathbf{c}^{(k)} \times \mathbf{c}^{(k+1)}}{||\mathbf{c}^{(k)} \times \mathbf{c}^{(k+1)}||_2}, \frac{  \mathbf{L}_{ij}}{ ||\mathbf{L}_{ij}||_2}\rangle \cdot \bar{\phi}^{(k,k+1)} \\
      \mathrm{where} \quad\mathbf{c}  & =\frac{\mathbf{L}_{ij}\times \mathbf{l}}{||\mathbf{L}_{ij}\times \mathbf{l}||_2} , \\
      \, \, \mathbf{c}^{(k)} \times \mathbf{c}^{(k+1)}  & = \frac{(\mathbf{L}_{ij}\times \mathbf{l}^{(k)})\times(\mathbf{L}_{ij}\times \mathbf{l}^{(k+1)})}{||\mathbf{L}_{ij}\times \mathbf{l}^{(k)}||_2||\mathbf{L}_{ij}\times \mathbf{l}^{(k+1)}||_2}, \\
      \, & = \frac{(\mathbf{L}_{ij} \cdot (\mathbf{l}^{(k)}\times \mathbf{l}^{(k+1)}))\mathbf{L}_{ij}}{||\mathbf{L}_{ij}\times \mathbf{l}^{(k)}||_2||\mathbf{L}_{ij}\times \mathbf{l}^{(k+1)}||_2}, \\
      \mathrm{thus} \, \,& \langle\frac{ \mathbf{c}^{(k)} \times \mathbf{c}^{(k+1)}}{||\mathbf{c}^{(k)} \times \mathbf{c}^{(k+1)}||_2}, \frac{  \mathbf{L}_{ij}}{ ||\mathbf{L}_{ij}||_2}\rangle  \\
      %= &  \langle\frac{(\mathbf{L}_{ij} \cdot (\mathbf{l}^{(k)}\times \mathbf{l}^{(k+1)}))\mathbf{L}_{ij} }{||\mathbf{c}^{(k)} \times \mathbf{c}^{(k+1)}||_2||\mathbf{L}_{ij}\times \mathbf{l}^{(k)}||_2||\mathbf{L}_{ij}\times \mathbf{l}^{(k+1)}||_2}, \frac{  \mathbf{L}_{ij}}{ ||\mathbf{L}_{ij}||_2}\rangle \\ 
      %= &  \frac{\mathbf{L}_{ij} \cdot (\mathbf{l}^{(k)}\times \mathbf{l}^{(k+1)}) }{||\mathbf{c}^{(k)} \times \mathbf{c}^{(k+1)}||_2||\mathbf{L}_{ij}\times \mathbf{l}^{(k)}||_2||\mathbf{L}_{ij}\times \mathbf{l}^{(k+1)}||_2} \\ 
      = &  \frac{\mathcal{R}(\mathbf{L}_{ij}) \cdot (\mathcal{R}(\mathbf{l}^{(k)}\times \mathbf{l}^{(k+1)})) }{||\mathbf{c}^{(k)} \times \mathbf{c}^{(k+1)}||_2||\mathbf{L}_{ij}\times \mathbf{l}^{(k)}||_2||\mathbf{L}_{ij}\times \mathbf{l}^{(k+1)}||_2} \\ 
      \varphi^{(k-1,k)} & = \arccos({\langle \mathbf{n}^{(k-1)} ,\mathbf{n}^{(k)} \rangle}) , \\
      = & = \arccos({\mathcal{R}(\langle \mathbf{n}^{(k-1)} ,\mathbf{n}^{(k)} \rangle})).
\end{split} 
\end{align*}
All extracted geometric features are invariant under rotation and translation transformations.
\end{proof}

\subsection{Proof of Theorem \ref{theo:theorem2}}\label{appendix:theorem2}
The proof of Theorem  \ref{theo:theorem2} is a consequence of the following \ref{lemma:1}.
\begin{lemma}\label{lemma:1} 
Given Cartesian coordinates of two end nodes $v_i$ and $v_j$, and one arbitrary point $p^{(k-1)}$ in a spatial curve $l_{ij}$, the Cartesian coordinate $p^{(k)}$ that form line segment $l^{(k)}$ with $p^{(k-1)}$ can be determined by the representation defined in Equation~\ref{eq:representation}.
\end{lemma}

\noindent Then we provide the proof for Theorem \ref{theo:theorem2}.
\begin{proof}
    As stated in Lemma \ref{lemma:1}, the Cartesian coordinates of a point $p^{(k)}$ can be derived from its two endpoints and their connected neighbor points $p^{(k-1)}$. Leveraging the connectivity of the spatial graph, we can iteratively compute the coordinates of a connected point based on the set of points with known coordinates. By initiating the process from any arbitrary point, we can determine the Cartesian coordinates for the entire spatial network.
\end{proof}
\begin{proof}
    Note that from Equation~\ref{eq:representation} we have followings:
    \begin{equation*}
    \begin{split}
         ||\mathbf{L}_{i,j}\times \mathbf{l}^{(k)}||_2 & = ||(\mathbf{p}_i-\mathbf{p}_j)\times (\mathbf{p}^{(k)}-\mathbf{p}^{(k-1)})||_2 = d^{k}d_{i,j}\sin\theta^{(k)},\\
         \mathbf{c}^{(k)} \times \mathbf{c}^{(k-1)} & = \frac{(\mathbf{L}_{i,j}\times (\mathbf{p}^{(k)}-\mathbf{p}_{j})) \times (\mathbf{L}_{i,j}\times (\mathbf{p}^{(k-1)}-\mathbf{p}_{j}))}{d_{i,j}^{2}d^{(k)}d^{(k-1)}\sin\theta^{(k)}\sin\theta^{(k-1)}}  \\
         & = \frac{( \mathbf{L}_{i,j} \cdot ((\mathbf{p}^{(k)}-\mathbf{p}_{j})\times (\mathbf{p}^{(k-1)}-\mathbf{p}_{j})))\mathbf{L}_{i,j}}{d_{i,j}^{2}d^{(k)}d^{(k-1)}\sin\theta^{(k)}\sin\theta^{(k-1)}} \\
    \end{split}
    \end{equation*}
    \begin{equation*}
    \begin{split}
         \langle\mathbf{c}^{(k)}, \mathbf{c}^{(k-1)}\rangle & =  \cos\bar{\phi}^{(k,k-1)} = \cos{\phi}^{(k,k-1)}, \\
         \frac{\langle \mathbf{c}^{(k)} \times \mathbf{c}^{(k-1)},  \mathbf{L}_{i,j}\rangle} {d_{i,j}\sin\bar{\phi}^{(k,k-1)}}  & =  \frac{ \mathbf{L}_{i,j} \cdot ((\mathbf{p}_j-\mathbf{p}^{(k)})\times (\mathbf{p}_j-\mathbf{p}^{(k-1)}))}{d_{i,j}d^{(k)}d^{(k-1)}\sin\theta^{(k)}\sin\theta^{(k-1)}\sin\bar{\phi}^{(k,k-1)}}. 
    \end{split}
    \end{equation*}
    Suppose there exists two different positions of point $p^{(k)}$ and $p^{(k)\prime}$ that satisfy Equation~\ref{eq:representation}, then it implies that
    \begin{equation*}
    \begin{split}
         \, & \frac{ \mathbf{L}_{i,j} \cdot ((\mathbf{p}_j-\mathbf{p}^{(k)})\times (\mathbf{p}_j-\mathbf{p}^{(k-1)}))}{d_{i,j}d^{(k)}d^{(k-1)}\sin\theta^{(k)}\sin\theta^{(k-1)}\sin\bar{\phi}^{(k,k-1)}} \\
         \, & = \frac{ \mathbf{L}_{i,j} \cdot ((\mathbf{p}_j-\mathbf{p}^{(k)\prime})\times (\mathbf{p}_j-\mathbf{p}^{(k-1)}))}{d_{i,j}d^{(k)}d^{(k-1)}\sin\theta^{(k)}\sin\theta^{(k-1)}\sin\bar{\phi}^{(k,k-1)}} \\
         0 & = (\mathbf{p}^{(k)} - \mathbf{p}^{(k)\prime}) \cdot ((\mathbf{p}_{j} - \mathbf{p}^{(k)})\times (\mathbf{p}_{j} - \mathbf{p}^{(k-1)})).
    \end{split}
    \end{equation*}
    Here $((\mathbf{p}_{j} - \mathbf{p}^{(k)})\times (\mathbf{p}_{j} - \mathbf{p}^{(k-1)}))$ is nonzero as long as $\mathbf{p}_{j}$, $\mathbf{p}^{(k)}$, $\mathbf{p}^{(k-1)}$ is non-colinear. Therefore, the above equation causes a contradiction which proves that $p^{(k)}$ can not have two different positions given the representation in Equation~\ref{eq:representation}.
\end{proof}

\subsection{Theoretical guarantee on approximation bound} \label{subsection:theory}
%It is also worth noticing that in our proposed model we represent the spatial path embedded in a manifold as a sequence of line segments and their embedded meshes to handle the continuous spatial data. However, since the mesh data approximates continuous manifold surfaces by many connected discrete triangle surfaces, it would have an approximation error from the ground truth spatial surface. More concretely, in our model, we calculate the spatial path representation in Equation~\ref{eq:representation} on each mesh unit. A key question is how different the calculated geometric features are from the ground truth, which is critical to the quality of learned representations. In the following, we further give a theoretical analysis of the approximation bound of few calculated spatial geometric features with respect to the size of meshes used. Specifically, we focus on the theoretical guarantee of angle $\varphi$, which describes the crucial relative orientation environment of the embedded mesh data.

It is crucial to note that our proposed model represents the spatial path embedded in a manifold as a sequence of line segments and their corresponding embedded meshes to handle continuous spatial data. However, since the mesh data approximates continuous manifold surfaces using discrete triangle surfaces, there exists an approximation error compared to the ground truth spatial surface. A key aspect is to understand the dissimilarity between the calculated geometric features and the ground truth, as this significantly impacts the quality of the learned representations. In the following analysis, we provide a theoretical examination of the approximation bound for calculated spatial geometric features concerning the size of the discretized meshes. Specifically, we focus on the theoretical guarantee of the angle $\varphi$, which characterizes the essential relative orientation environment of the embedded mesh data.

To facilitate our analysis, we introduce the following notations: the empty ball size $ebs(p)$ denotes the radius of the smallest medial ball tangent to $M$ at point $p$, and the local feature size $lfs(p)$ represents the distance between point $p$ and the nearest point on $M$. Additionally, $\vert pq\vert$ denotes the distance between point $p$ and point $q$, $n_p$ denotes an outward-directed vector normal to $M$ at point $p$, and $N^f_p$ represents a vector normal to any face $f$ at point $p$. The angle between two vectors $n_p$ and $n_q$ is denoted as $\angle (n_p,n_q)$.
%\indent Before presenting the angle bound between any pair of faces, we introduce some notations: the empty ball size $ebs(p)$ is the radius of the smallest medial ball tangent to $M$ at point $p$; the local feature size $lfs(p)$ is the distance between point $p$ and the nearest point on $M$; $\vert pq\vert$ means the distance between point $p$ and point $q$; $n_p$ means an outward-directed vector normal to $M$ at point $p$. $N^f_p$ is a vector normal to any face $f$ at point $p$. $\angle (n_p,n_q)$ is the angle between two vectors $n_p$ and $n_q$. %The following definitions are also required for our theoretical analysis:
\begin{definition}[Plane Angle]
For triangle face $f^{(i,j)}_{k}$, the plane angle at vertex $v^{(i,j)}_{k}$ denotes its associated triangle angle.%Because a face is a triangle, then for any face $f^{(i,j)}_{k}$, a plane angle of $f^{(i,j)}_{k}$ at vertex $v^{(i,j)}_{k}$ is the triangle angle associated with vertex $v^{(i,j)}_{k}$. 
\end{definition}
\begin{definition}[Circumradius]
The diametric ball $B^{(i,j)}_{k}$ of a face $f^{(i,j)}_{k}$ is defined as the smallest closed ball that contains $f^{(i,j)}_{k}$ and has all its vertices lying on the boundary of $B^{(i,j)}_{k}$. The circumcenter and circumradius correspond to the center and radius of $B^{(i,j)}_{k}$, respectively.%For any face $f^{(i,j)}_{k}$, its  diametric ball $B^{(i,j)}_{k}$ is the smallest closed ball such that $f^{(i,j)}_{k}\subset B^{(i,j)}_{k}$ and all vertices of $f^{(i,j)}_{k}$ lie in the boundary of $B^{(i,j)}_{k}$. circumcenter and circumradius are the center and radius of $B^{(i,j)}_{k}$.
\end{definition}
%\indent Based on the above notations and definitions, we have the following angle bound:
\begin{theorem} [Angle Bound]
 Assume $M$ is bounded and smooth. Let $f^{(i_1,j_1)}_{k_1}$ and $f^{(i_2,j_2)}_{k_2}$ be two faces whose vertices lie on $M$, while $R^{(i_1,j_1)}_{k_1}$ and $R^{(i_2,j_2)}_{k_2}$ be circumradii (i.e. plural form of circumradius) of $f^{(i_1,j_1)}_{k_1}$ and $f^{(i_2,j_2)}_{k_2}$, respectively. Let $\phi ^{(i_1,j_1)}_{k_1}$ and $\phi ^{(i_2,j_2)}_{k_2}$ be plane angles of $f^{(i_1,j_1)}_{k_1}$ and $f^{(i_2,j_2)}_{k_2}$ at $v^{(i_1,j_1)}_{k_1}$ and $v^{(i_2,j_2)}_{k_2}$, respectively. Let $\delta=\vert v^{(i_1,j_1)}_{k_1}v^{(i_2,j_2)}_{k_2}\vert/lfs(v^{(i_1,j_1)}_{k_1})$. If $\delta<\sqrt{4\sqrt{5}-8}$, then
\begin{align}
       \nonumber &\angle (N^{f^{(i_1,j_1)}_{k_1}}_{v^{(i_1,j_1)}_{k_1}},N^{f^{(i_2,j_2)}_{k_2}}_{v^{(i_2,j_2)}_{k_2}}) \leqslant \arccos(1-\frac{\delta^2}{2\sqrt{1-\delta^2}})\\\nonumber&+\arcsin(\frac{R^{(i_1,j_1)}_{k_1}}{ebs(v^{(i_1,j_1)}_{k_1})}\max(\cot(\frac{\phi ^{(i_1,j_1)}_{k_1}}{2}),1))\\&+\arcsin(\frac{R^{(i_2,j_2)}_{k_2}}{ebs(v^{(i_2,j_2)}_{k_2})}\max(\cot(\frac{\phi ^{(i_2,j_2)}_{k_2}}{2}),1))
       \label{eq: angle bound}
\end{align}
where $\angle (N^{f^{(i_1,j_1)}_{k_1}}_{v^{(i_1,j_1)}_{k_1}},N^{f^{(i_2,j_2)}_{k_2}}_{v^{(i_2,j_2)}_{k_2}})$ is the angle between two vectors $N^{f^{(i_1,j_1)}_{k_1}}_{v^{(i_1,j_1)}_{k_1}}$ and $N^{f^{(i_2,j_2)}_{k_2}}_{v^{(i_2,j_2)}_{k_2}}$ normal to $f^{(i_1,j_1)}_{k_1}$ and $f^{(i_2,j_2)}_{k_2}$ at vertices $v^{(i_1,j_1)}_{k_1}$ and $v^{(i_2,j_2)}_{k_2}$, respectively.
\label{theo: angle bound}
\end{theorem}

Before proving Theorem \ref{theo: angle bound}, we first prove that the angle satisfies the triangle inequality.
\begin{lemma}\cite{angle_triangle_inequality}
For any three vectors $\vec{x},\vec{y},\vec{z}$, we have $\angle(\vec{x},\vec{y})\leq \angle(\vec{x},\vec{z})+\angle(\vec{z},\vec{y})$.
\label{lemma:angle}
\end{lemma}
\begin{proof}
\indent Without loss of generality, we assume that $\vec{x},\vec{y}$ and $\vec{z}$ are unit vectors. Let 
\begin{align*}
G=
    \begin{bmatrix}
    1&\langle \vec{x},\vec{y}\rangle&\langle \vec{x},\vec{z}\rangle\\\langle\vec{y},\vec{x}\rangle&1&\langle \vec{y},\vec{z}\rangle\\\langle \vec{z},\vec{x}\rangle&\langle \vec{z},\vec{y}\rangle&1\\
    \end{bmatrix}
\end{align*}
Note that $G$ is a Gramian matrix, and hence positive-semidefinite. Thus $\det G\geq0$, we have
\begin{align*}
    1+2\langle \vec{x},\vec{y}\rangle\langle \vec{y},\vec{z}\rangle\langle \vec{z},\vec{x}\rangle-\vert \langle\vec{x},\vec{y}\rangle\vert^2-\vert \langle \vec{y},\vec{z}\rangle\vert^2-\vert \langle \vec{z},\vec{x}\rangle\vert^2\geq 0.
\end{align*}
This is equivalent to
\begin{align*}
    (1-\vert\langle \vec{x},\vec{y}\rangle\vert^2)(1-\vert\langle \vec{y},\vec{z}\rangle\vert^2)\geq\vert \langle \vec{x},\vec{y}\rangle\langle \vec{y},\vec{z}\rangle-\langle \vec{z},\vec{x}\rangle\vert^2.
\end{align*}
by expanding both sides and canceling the common term$\vert \langle \vec{x},\vec{y}\rangle\vert^2\vert \langle \vec{y},\vec{z}\rangle\vert^2$. Taking the square root of both sides, we get
\begin{align*}
    \vert  \langle \vec{x},\vec{y}\rangle \langle \vec{y},\vec{z}\rangle\vert -\vert  \langle \vec{z},\vec{x}\rangle\vert &\leq  \vert  \langle \vec{x},\vec{y}\rangle \langle \vec{y},\vec{z}\rangle -  \langle \vec{z},\vec{x}\rangle\vert\\&\leq \sqrt{1-\vert  \langle \vec{x},\vec{y}\rangle\vert^2}\sqrt{1-\vert  \langle \vec{y},\vec{z}\rangle\vert^2},
\end{align*}
thus
\begin{align*}
    \vert \langle\vec{z},\vec{x}\rangle\vert\geq \vert  \langle \vec{x},\vec{y}\rangle \langle \vec{y},\vec{z}\rangle\vert-\langle \vec{z},\vec{x}\rangle\vert\geq \sqrt{1-\vert  \langle \vec{x},\vec{y}\rangle\vert^2}\sqrt{1-\vert  \langle \vec{y},\vec{z}\rangle\vert^2}
\end{align*}
Note that $\cos\angle (\vec{x},\vec{y})=\vert \langle\vec{x},\vec{y}\rangle\vert$ holds for any unit vectors $\vec{x}$, $\vec{y}$, so this inequality becomes
\begin{align*}
    \cos\angle (\vec{x},\vec{z})&\geq \cos\angle (\vec{x},\vec{y})\cos\angle (\vec{y},\vec{z}) \\ &-\sqrt{1-\cos  \langle \vec{x},\vec{y}\rangle^2}\sqrt{1-  \cos \langle\vec{y},\vec{z}\rangle^2}\\&=\cos\angle (\vec{x},\vec{y})\cos\angle (\vec{y},\vec{z})-\sin\angle (\vec{x},\vec{y})\sin\angle (\vec{y},\vec{z})\\&=\cos(\angle (\vec{x},\vec{y})+\angle (\vec{y},\vec{z})) 
\end{align*}
here $\vert\sin\angle (\vec{x},\vec{y})\vert=\sin\angle (\vec{x},\vec{y})$ and $\vert\sin\angle (\vec{y},\vec{z})\vert=\sin\angle (\vec{y},\vec{z})$ because $\angle (\vec{x},\vec{y})\in[0,\pi],\angle (\vec{y},\vec{z})\in[0,\pi]$.
Because $\cos(\cdot)$ is strictly monotonically decreasing on $[0,\pi]$, we have
\begin{align*}
    \angle (\vec{x},\vec{z})\leq \angle (\vec{x},\vec{y})+\angle (\vec{y},\vec{z}).
\end{align*}
This proof is completed.
\end{proof}
\indent Now we prove Theorem \ref{theo: angle bound} as follows:
\begin{proof}
By applying Lemma \ref{lemma:angle}, we have
\begin{align*}
\angle (N^{f^{(i_1,j_1)}_{k_1}}_{v^{(i_1,j_1)}_{k_1}},N^{f^{(i_2,j_2)}_{k_2}}_{v^{(i_2,j_2)}_{k_2}})&\leq \angle (N^{f^{(i_1,j_1)}_{k_1}}_{v^{(i_1,j_1)}_{k_1}},N^M_{v^{(i_1,j_1)}_{k_1}})+ \\ \angle (N^M_{v^{(i_1,j_1)}_{k_1}}), & (N^M_{v^{(i_2,j_2)}_{k_2}}) + \angle (N^M_{v^{(i_2,j_2)}_{k_2}},N^{f^{(i_2,j_2)}_{k_2}}_{v^{(i_2,j_2)}_{k_2}}).   
\end{align*}
By Lemma 2 in \cite{khoury2019approximation}, we have
\begin{align*}
 &\angle (N^{f^{(i_1,j_1)}_{k_1}}_{v^{(i_1,j_1)}_{k_1}},N^M_{v^{(i_1,j_1)}_{k_1}})\leq \\
 &\arcsin(\frac{R^{(i_1,j_1)}_{k_1}}{ebs(v^{(i_1,j_1)}_{k_1})}\max(\cot(\frac{\phi ^{(i_1,j_1)}_{k_1}}{2}),1)),\\
 & \angle (N^M_{v^{(i_2,j_2)}_{k_2}},N^{f^{(i_2,j_2)}_{k_2}}_{v^{(i_2,j_2)}_{k_2}})\leq \\
 & \arcsin(\frac{R^{(i_2,j_2)}_{k_2}}{ebs(v^{(i_2,j_2)}_{k_2})}\max(\cot(\frac{\phi ^{(i_2,j_2)}_{k_2}}{2}),1)).
\end{align*}
By Lemma 4 in \cite{khoury2019approximation}, we have
\begin{align*}
 \angle (N^M_{v^{(i_1,j_1)}_{k_1}}), (N^M_{v^{(i_2,j_2)}_{k_2}})   \leq \arccos(1-\frac{\delta^2}{2\sqrt{1-\delta^2}})
\end{align*}
We sum up all these inequalities, and Theorem \ref{theo: angle bound} is proven.
\end{proof}

\section{Additional Experimental Settings and Results}

\subsubsection{Rotation and translation invariant test} \label{appendix:invariant}
\begin{figure}[t]
    \centering
        \includegraphics[width=0.5\textwidth]{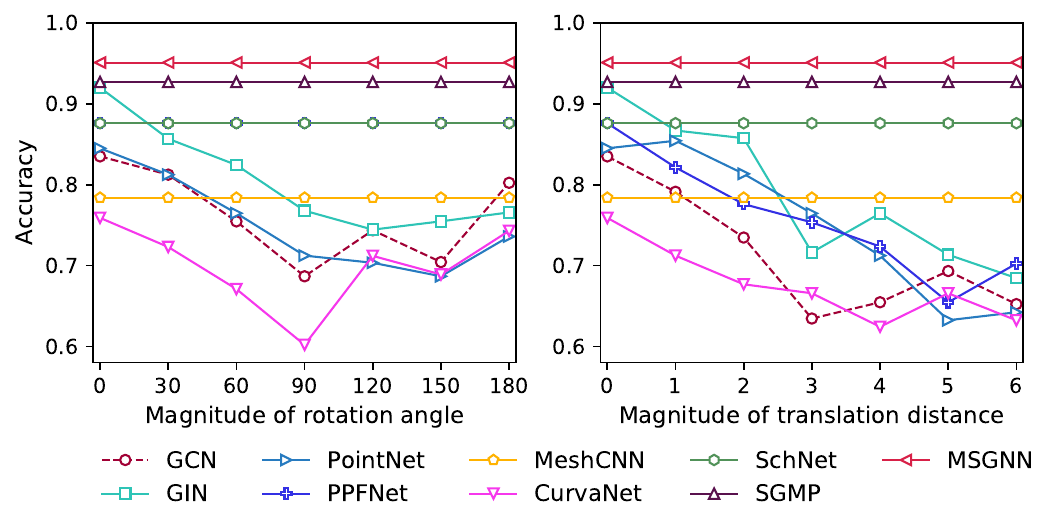}    
        %\captionsetup{font=small}
        \vspace{-5mm}
        \captionsetup{width=1.\linewidth}
        \caption{Robustness test on rotation and translation invariance by augmenting the data on the test set. The x-axis corresponds to the magnitude of the rotation angle (left) and translation distance (right) while y-axis shows the accuracy scores.}
        \label{fig:roto_translation} 
\end{figure}

Similar to previous work~\cite{fuchs2020se,zhang2021representation}, we also measure the robustness of rotation and translation by uniformly adding translation and rotation transformations to the input Cartesian coordinates. Due to space limitations, here we only report the accuracy results of the classification task on the brain dataset HCP, and the results are similar on all datasets. According to Figure~\ref{fig:roto_translation}, we can notice that the performance of our proposed model remains unchanged under all translation and rotation transformations. SchNet, SGMP, and MeshCNN can also achieve invariance under transformations because they also use only rotation- and translation-invariant geometric features in their models. PPFNet can be invariant under rotation transformation, but not translation transformation because it preserves the origin in the model. This experiment validates the importance of applying a rotation- and translation-invariant model, as we can observe that the performance of models without theoretical guarantees drops significantly when adding rotation and translation transformations. While our MSGNN model's performance is consistent with the theoretical guarantees that the learned representations are rigorously invariant under rotation and translation transformations.

\subsection{Detailed datasets introduction.}\label{appendix:dataset}
\noindent (1) \textbf{HCP brain networks.} 
The cerebral cortex, the outer layer of neural tissue in the human brain, exhibits a complex geometric surface characterized by numerous folds and grooves~\cite{glasser2016multi, hamalainen1993magnetoencephalography}. This irregular structure, representing a two-dimensional manifold surface, plays a vital role in understanding the connectivity patterns of neurons. Our objective is to classify the activity states of subjects based on processed functional connectivity (FC) networks derived from the cortical manifold environment. This task is crucial for uncovering the underlying mechanisms of neural connectivity~\cite{leonardi2013principal}. The data used in our study is obtained from the Human Connectome Project (HCP), specifically Magnetic Resonance Imaging (MRI) data~\cite{van2013wu}. Each data sample is associated with an activity state (e.g., rest, gamble) as the target for prediction. The cerebral cortex's geometric shape is divided into 68 predefined regions of interest (ROIs) using probabilistic tracking on diffusion MRI data, employing the Probtrackx tool from the FMRIB Software Library~\cite{jenkinson2012fsl}. Functional connectivity (FC) is computed as the Pearson correlation between the blood oxygen level-dependent time series of two ROIs, extracted from resting-state functional MRI data. To construct brain networks, a threshold is applied to the FC values to filter out highly correlated edges. The spatial trajectories of interest are defined as geodesic paths between the centers of the ROIs.

\noindent (2) \textbf{Air transportation networks.} Forecasting the future structure of the air transport network has important implications for the economic integration and development prospects of regions and countries. Link prediction aims to predict whether two airports will be connected by direct flights in the future development stage of flight network~\cite{lu2011link}. Specifically, we adopt two publicly available flight networks \textit{Flight-NA} and \textit{Flight-GL}, where \textit{Flight-NA} contains 456 airports and 71,959 airlines in the North America~\cite{frey2007clustering} and \textit{Flight-GL}~\cite{Jia-2019-CP} contains 3,214 airports and 66,771 airlines spanning the globe. The earth surface is considered as the manifold to include the curved airline trajectory.

\noindent (3) \textbf{3D shapes classification.} We further conduct experiments on classifying 3D shapes in two datasets \textit{SHREC} and \textit{FAUST}, where \textit{SHREC} contains watertight triangle meshes with 30 classes and \textit{FAUST} contains scans of human poses in a wide range. We follow previous studies~\cite{hanocka2019meshcnn} to sample a lower resolution ($\sim$500 faces) from a higher solution. The vertices of the lower-resolution triangle tessellation are then treated as the nodes and their geodesic trajectories are treated as the spatial curves.

\subsection{Comparison methods details.}
\noindent (1) \textbf{GCN}~\cite{kipf2016semi} is a commonly used GNN model by using a localized first-order approximation of spectral graph convolutions;

\noindent (2) \textbf{GIN}~\cite{xu2018powerful} is a GNN model with provably powerful discriminating power among the class of 1-order GNNs;

\noindent (3) \textbf{PointNet}~\cite{qi2017pointnet} is a general deep learning method for representing the global 3D shape as a set of points in Euclidean space;

\noindent (4) \textbf{PPFNet}~\cite{deng2018ppfnet} is a spatial deep learning framework to learn a globally aware 3D descriptor by constructing point pair features within the local vicinity;

\noindent (5) \textbf{MeshCNN}~\cite{hanocka2019meshcnn} is a deep learning method to learn the representations for mesh data. It applies convolutions on mesh edges and the four edges of their incident triangles; 

\noindent (6) \textbf{CurvaNet}~\cite{he2020curvanet} is a deep learning method for mesh data by further treating the curvature value of each mesh unit with respect to its surrounding units;

\noindent (7) \textbf{SchNet}~\cite{schutt2017schnet} is a deep learning model designed for spatial networks in Euclidean space. It utilizes a continuous filter function on the distances between nodes and their first-order neighbors;

\noindent (8) \textbf{SGMP}~\cite{zhang2021representation} is a deep learning method for spatial networks in Euclidean space. It further includes the directional information by aggregating the length three path messages based on a physical representation of distances, angles, and torsions. %With the extracted geometric features, their methods can guarantee the preservation of geometric shape information in Euclidean space.

\end{document}